\documentclass[conference]{IEEEtran}
\makeatletter
\def\ps@headings{%
\def\@oddhead{\mbox{}\scriptsize\rightmark \hfil \thepage}%
\def\@evenhead{\scriptsize\thepage \hfil \leftmark\mbox{}}%
\def\@oddfoot{}%
\def\@evenfoot{}}
\makeatother
\usepackage{cite}
\usepackage{amsmath,amssymb,amsfonts}
\usepackage{graphicx}
\usepackage{textcomp}

\usepackage{subfigure}
\usepackage{algorithmic}

\usepackage{xcolor}
\usepackage{caption}
\usepackage{wrapfig,lipsum,booktabs}
\usepackage[export]{adjustbox}
\definecolor{red}{rgb}{2,0,0}
\usepackage{graphicx}
\ifCLASSOPTIONcompsoc
    \usepackage[caption=false, font=normalsize, labelfont=sf, textfont=sf]{subfig}
\else
\usepackage[caption=false, font=footnotesize]{subfig}
\fi
\def\BibTeX{{\rm B\kern-.05em{\sc i\kern-.025em b}\kern-.08em
    T\kern-.1667em\lower.7ex\hbox{E}\kern-.125emX}}

\IEEEoverridecommandlockouts

\DeclareMathAlphabet{\mathpzc}{OT1}{pzc}{m}{it}
\newtheorem{definition}{Definition}

\usepackage[ruled, vlined, linesnumbered, commentsnumbered, longend]{algorithm2e}

\SetCommentSty{mycommfont}
\usepackage[export]{adjustbox}
\usepackage{booktabs}
\begin{document}
    
\title{\Large \uppercase{Hiding in Plain Sight: Differential Privacy Noise Exploitation for Evasion-resilient Localized Poisoning Attacks in Multiagent Reinforcement Learning}}

\author{
\IEEEauthorblockN{\textbf{\uppercase{Md Tamjid Hossain}}\IEEEauthorrefmark{1}, \textbf{\uppercase{Hung La}}\IEEEauthorrefmark{2}}

\IEEEauthorblockA{Advanced Robotics and Automation (ARA) Laboratory\\ University of Nevada, Reno, NV, USA
}

\IEEEauthorblockA{
E-mail: 
\IEEEauthorrefmark{1}mdtamjidh@nevada.unr.edu, \IEEEauthorrefmark{2}hla@unr.edu}

\thanks{This work was partially funded by the U.S. National Science Foundation (NSF) under grant NSF-CAREER: 1846513.  The views, opinions, findings, and conclusions reflected in this publication are solely those of the authors and do not represent the official policy or position of the NSF.}
\thanks{*Source code of the task and the computational model behind the setup available at {\tt \small https://github.com/aralab-unr/PeLPA}}%

}

\IEEEaftertitletext{\vspace{-1.0\baselineskip}}

\maketitle

\begin{abstract}
Lately, differential privacy (DP) has been introduced in cooperative multiagent reinforcement learning (CMARL) to safeguard the agents' privacy against adversarial inference during knowledge sharing. Nevertheless, we argue that the noise introduced by DP mechanisms may inadvertently give rise to a novel poisoning threat, specifically in the context of private knowledge sharing during CMARL, which remains unexplored in the literature. To address this shortcoming, we present an adaptive, privacy-exploiting, and evasion-resilient localized poisoning attack (PeLPA) that capitalizes on the inherent DP-noise to circumvent anomaly detection systems and hinder the optimal convergence of the CMARL model. We rigorously evaluate our proposed PeLPA attack in diverse environments, encompassing both non-adversarial and multiple-adversarial contexts. Our findings reveal that, in a medium-scale environment, the PeLPA attack with attacker ratios of $20\%$ and $40\%$ can lead to an increase in average \textit{steps to goal} by $50.69\%$ and $64.41\%$, respectively. Furthermore, under similar conditions, PeLPA can result in a $1.4$x and $1.6$x computational time increase in optimal reward attainment and a $1.18$x and $1.38$x slower convergence for attacker ratios of $20\%$ and $40\%$, respectively.
\end{abstract}

\begin{IEEEkeywords}
Differential Privacy, Adversarial Learning, Poisoning Attacks, Cooperative Multiagent Reinforcement Learning
\end{IEEEkeywords}

\section{Introduction} \label{s:Introduction}

Cooperative multiagent reinforcement learning (CMARL) has been acknowledged for its proficiency in orchestrating complex tasks, such as automated robotic swarming and distributed power system optimization, through multi-agent collaboration \cite{Zhou2022,da2017simultaneously,le2017coordinated}. However, the inherent nature of data sharing in CMARL can trigger potential privacy infringements, as the shared experiences often encompass sensitive data \cite{li2022privacy,zou2020privacy}. To combat this, differential privacy (DP) mechanisms \cite{dwork2006}, which employ stochastic noise addition to obfuscate sensitive data, are posited as effective countermeasures \cite{li2022privacy,ye2022differential, Abahussein2023,ye2022one}.

Yet, we conjecture that adversaries could exploit DP's noise-adding mechanism to craft their own malicious noise in CMARL, thereby degrading the learning efficacy while remaining undetected by hiding behind the DP-noise, leading to catastrophic implications in sectors like robotics, cyber-physical systems, automotive industries, etc. \cite{scheikl2021cooperative,la2014multirobot,prasad2019multi,kiran2021deep}. For example, false advising with DP-exploited misleading knowledge from advisor cars in autonomous driving may make lane-changing ambiguous and lead to severe road accidents. Contemporary state-of-the-art (SOTA) poisoning attacks typically focus on voluminous malicious data injection, which is prone to detection, leaving the creation of subtle, stealthy adversarial instances as a formidable challenge \cite{cao2021data,fang2020local, mohammadi2023implicit,cheu2021manipulation}.

Addressing this challenge, our research proposes a novel adversarial model tailored for CMARL that exploits DP-induced noise to facilitate stealthy, localized poisoning attacks \cite{Hossain2021Desmp,giraldo2020adversarial,hossain2022adversarial}. To our knowledge, this is the first investigation into DP-noise exploitation for conducting local poisoning attacks while evading detection in CMARL. Our contributions are:

\begin{itemize}
\item Uncovering the susceptibility of DP mechanisms to adversarial poisoning attacks, illustrating how adversaries can adaptively perturb knowledge to remain undetected.
\item Proposing a novel privacy-exploiting local poisoning attack (PeLPA), contrasting general poisoning attacks that overlook the importance of attack stealthiness.
\item Experimentally evaluating the potential ramifications of DP-exploited stealthy attacks in safety-critical sectors.
\end{itemize}

The terms knowledge', experience', advice', and Q-value' are used interchangeably throughout the paper.
\section{Related Works} \label{s:litReview}
In this section, we address the SOTA poisoning techniques.
\subsection{Application of Differential Privacy for Knowledge Sharing}

DP, a prominent method for privacy preservation, has been extensively employed in private knowledge sharing within the realm of CMARL \cite{ye2022differential,wang2019privacy,li2022privacy, Wei2022, Abahussein2023,ye2022one}. The scope of its application includes DP-guided Q-learning models to maintain the privacy of reward data \cite{wang2019privacy}, privacy-centric multi-agent frameworks leveraging federated learning (FL) and DP to obstruct illegitimate access to data statistics \cite{li2022privacy}, and harnessing ($\beta, \phi$)-DP to counteract offloading preference inference attacks in vehicular ad-hoc networks (VANET) \cite{Wei2022}. Apart from protecting user information during private knowledge sharing, DP has also been proposed for differential advising. In particular, Ye et al. \cite{ye2022differential} propose a DP-based advising method for CMARL that enables agents to use the advice in a state even if the advice is created in a slightly different state. \textit{Nevertheless, they overlook the susceptibility of DP to poisoning attacks during knowledge sharing \cite{ye2022differential, Abahussein2023,ye2022one}.}

\subsection{Poisoning Attacks in Cooperative Multiagent Learning} The infiltration of poisoning attacks in CMARL, which can alter training datasets and consequently disrupt learning outcomes, is a pertinent research concern \cite{figura2021adversarial,fang2020local, Xie2022,mohammadi2023implicit}. Research has delved into scenarios where adversarial agents can manipulate network-wide policies \cite{figura2021adversarial}, scrutinized targeted poisoning attacks in dual-agent frameworks where one agent’s policy is modified \cite{mohammadi2023implicit}, and investigated the implications of soft actor-critic algorithms in CMARL for executing poisoning attacks \cite{Xie2022}. For instance, Figura et al. \cite{figura2021adversarial} demonstrate that an adversarial agent can persuade all other agents in the network to implement policies that optimize its desired objective. Another approach for performing poisoning attacks by any malicious advisor in multiagent Q-learning as demonstrated in \cite{hossain2023BRNES}, is to shuffle the Q-values for all actions corresponding to the requested state
and inject false noise that is similar to the maximum reward using reward poisoning method. \textit{However, the ramifications of these SOTA poisoning techniques against anomaly detection and privacy-preserving knowledge-sharing technologies remain largely unexplored. Our work endeavors to model a DP noise-exploiting poisoning attack that remains resilient to detection algorithms.}

\subsection{Differential Privacy Exploitation Techniques} Another domain of interest focuses on the possible exploitation of DP in classification challenges, even though it does not necessarily concentrate on adversarial onslaughts on CMARL algorithms \cite{giraldo2017security_2, giraldo2020adversarial, hossain2021privacy,cao2021data,cheu2021manipulation,hossain2022adversarial, Hossain2021Desmp}. This research trajectory involves the systemic degradation of utility by exploiting DP noise \cite{giraldo2020adversarial}, gauging the impact of DP manipulation in smart grid networks \cite{hossain2021privacy}, and designing stealthy model poisoning attacks on an FL model \cite{Hossain2021Desmp, hossain2022adversarial}. Similarly, \cite{hossain2021privacy} investigates the impact of DP exploitation in a smart grid network and introduces a correlation among DP parameters to enable the system designer to calibrate the privacy level and reduce the attack surface. To examine the effect of DP-exploiting attacks on an FL model, \cite{Hossain2021Desmp} proposes a stealthy model poisoning attack leveraging DP noise added to ensure privacy. They improve their attack technique in \cite{hossain2022adversarial}, investigating how the degree of model poisoning can be adjusted dynamically through episodic loss memorization in FL and demonstrating how their attack can evade some SOTA defense techniques, such as norm, accuracy, and mix detection. \textit{However, these attack models face constraints in multi-agent environments or decentralized CMARL platforms.} Contrarily, Cao et al. \cite{cao2021data} propose an attack on the Local Differential Privacy (LDP) protocol by introducing fraudulent users. \textit{Our research, however, targets legitimate yet compromised users infusing false noise into shared data, also aiming to dodge anomaly detectors - a critical objective for a successful attack.}
\section{Local Differentially Private Cooperative Multiagent Reinforcement Learning}
\label{s:DP_CMARL}

We present a local differentially private CMARL (LDP-CMARL) framework akin to the one adopted in \cite{hossain2023BRNES}. However, for demonstration simplicity, instead of a generalized randomized response (GRR) technique, we leverage a Bounded Laplace (BLP) mechanism \cite{Neera2023} to model our LDP framework that also achieves the same $\varepsilon$-LDP guarantee.

\subsection{Cooperative Multiagent Reinforcement Learning (CMARL)}
\label{subsec:CMARL} \textbf{Environment model.} Our research formalizes a cooperative reinforcement learning context with a Markov game $\mathcal{M} =(\mathpzc{N}, \mathpzc{S}, \mathpzc{A}, \Phi, \Gamma, \mathpzc{T})$ incorporating $\mathpzc{N}$ robots navigating an environment $\mathbb{E}$ of dimensions height ($\mathpzc{H}$) and width ($\mathpzc{W}$) towards a goal $\mathpzc{G}$. It introduces obstacles $\mathpzc{O}$ and freeway $\mathpzc{F}$ with corresponding reward penalties and incentives, $\phi_{\mathpzc{O}}$ and $\phi_{\mathpzc{F}}$. Dynamic obstacle positioning adds complexity to learning, which concludes when the first agent reaches $\mathpzc{G}$.

\textbf{Learning objectives.} Agent $p_i$'s objective is to take the fewest steps, $\Pi$ to reach $\mathpzc{G}$, collect $\phi_f$, avoid hitting $o_x \in \mathpzc{O}$, and earn as much as rewards, $\phi_{\mathpzc{F},\mathpzc{G}}$. In short, the objectives can be formalized as
\begin{equation}
\begin{alignedat}{2}
(a)&\;\Pi{p_i} = \underset{\mathcal{M}}{min}\;\Pi\\
(b)&\;\phi_{p_i} = \phi_{\mathpzc{F}} + \phi_{\mathpzc{G}} + \left[\phi_{\mathpzc{O}} = 0\right]\\
(c)&\;\lVert (x_{p_i}, y_{p_i})- (x_{\mathpzc{G}}, y_{\mathpzc{G}}) \rVert = 0
 \end{alignedat}
 \forall \phi_{\mathpzc{G}, \mathpzc{F}, \mathpzc{O}} \in \Phi \text{ and } \phi_{\mathpzc{G}} > \phi_{\mathpzc{F}}
 \label{eqn:objective1}
\end{equation}

where $(x_{p_i}, y_{p_i})$, and $(x_{\mathpzc{G}}, y_{\mathpzc{G}})$ are $p_i$'s and $\mathpzc{G}$'s positions.

\subsection{Integrating Local Differential Privacy (LDP) in CMARL}
LDP protocols encapsulate two main stages: perturbation and aggregation. The Q-values domain, denoted as $\mathbb{Q} = \left[q\right]$, undergoes local perturbation before being relayed to the advisee, $p_i$, ensuring $p_i$'s inability to infer the original Q-value of the advisor, $p_k$. The aggregation phase facilitates $p_i$'s estimation of optimal advice utilizing the perturbed values received from all $p_k$, with perturbation function for Q-values of all actions, $a$ in state $s$ represented as $P(Q(s))$. Following the definition of $\varepsilon$-LDP \cite{cao2021data}, a protocol achieving LDP must ensure the probabilistic resemblance between any pair of perturbed Q-values.

LDP offers plausible deniability to $p_k$, restraining $p_i$ from determining the origin of the output confidently. This ambiguity is regulated by the privacy budget, $\varepsilon$ \cite{dwork2006}. To actualize $(\varepsilon, 0)$-DP, the Laplace mechanism, a noise-addition technique, is applied as follows \cite{dwork2006}:

\begin{equation}
    \mathpzc{M}(D) = f(D) + \eta \sim \mathcal{N}(0, b)
\end{equation}

where the added
noise, $\eta$ is drawn from a zero-mean Laplace distribution with
scale parameter, $b \geq \frac{\Delta}{\varepsilon}$. Here, $\Delta$ denotes the sensitivity of the query function. Nonetheless, the same Laplace mechanism that satisfies $(\varepsilon,0)$-DP,  can be deployed in a distributed fashion for achieving $\varepsilon$-LDP \cite{wang2020comprehensive, Neera2023}, by integrating randomized Laplace noise into each state-action pair's Q-values of an advisor. We leverage the higher noise sensitivity offered by the Laplace mechanism to attain stronger privacy protection as compared to Gaussian or Exponential mechanism. The advisee, $p_i$ computes the average value from all the noisy Q-values \cite{wang2020comprehensive}. We utilize the following BLP technique for input perturbation \cite{Neera2023}: 

\begin{definition}[Bounded Laplace Mechanism (BLP)]
   Given an input $q \in \left[l, u\right] \subset \mathbb{R}$, and scale $b>0$, the BLP technique, $\mathpzc{M}: \Omega \rightarrow \left[l, u\right]$ over output $\bar{q}$ can be represented by the following conditional probability density function (pdf):
   
   \begin{equation}
       f\mathpzc{M}(\bar{q}) = 
       \begin{cases}
       0 & \text{ if } \bar{q} \notin \left[l,u\right]\\
       \frac{1}{C_q} \frac{1}{2b}e^{-\frac{\lvert \bar{q}- q\rvert}{b}} & \text{ if } \bar{q} \in \left[l,u\right]
       \end{cases}
   \end{equation}
\end{definition}
where $l$ and $u$ are the lower and upper range, and $C_q = \int_{l}^{u} \frac{1}{2b}e^{-\frac{|\bar{q}- q|}{b}} \,d\bar{q}$ is a normalization constant. The proof and further details can be found in \cite{Neera2023}. BLP constrains noise sampling within a predefined range, avoiding values that may detriment learning performance. Hence, the sensitivity of the combined LDP mechanism is $\Delta = \lvert u-l\rvert$. Similar to \cite{ye2022differential}, within our LDP-CMARL framework, the sensitivity $\Delta$ needs to be calculated carefully. The LDP-CMARL framework training stages utilizing the BLP mechanism are outlined in Algorithm~\ref{algo:DP_CMARL}. During advice request dispatch, $p_i$ specifies a neighbor zone, $\mathpzc{Z}$, and sends advice requests only to advisors within $\mathpzc{Z}$. Both $p_i$ and $p_k$ calculate their advice requesting ($\varrho_{p_i}$) and advice giving ($\varrho_{p_k}$) probabilities as per \cite{ye2022differential}. After receiving advice from the neighbors, $p_i$ aggregates all the advice following a weighted linear aggregation technique, controlled by a predefined weight parameter, $w$ \cite{hossain2023BRNES}. Then, $p_i$ selects and executes an optimal action followed by a final Q-table update.

\setlength{\textfloatsep}{0pt}% 
\begin{algorithm}[!t]
    \SetKwFunction{LDP}{LDP}
    \SetKwProg{Fn}{Function}{:}{}
    \SetKwInOut{KwIn}{Input}
    \SetKwInOut{KwOut}{Output}

    \KwIn{$\mathbb{E}, \mathpzc{N}, \mathpzc{A}, \mathpzc{S}, \Phi \rightarrow (l, u)$}
    \KwOut{Trained LDP-CMARL model}

    Initialize Q-table, set $\varepsilon, \alpha, \Gamma$, and compute $b =\frac{\alpha\lvert u-l\rvert}{\varepsilon}$

    \For{each agent, $p_i \in \mathpzc{N}$}{
        \For{each episode}{
            Initialize state, $s$
            \For{each state}{
                Send advice request to $p_k$ in $\mathpzc{Z}$ with $\varrho_{p_i}$\\
                Receive LDP-advice, $\LDP(s, \varepsilon, b)\rightarrow \bar{Q}_{p_i}(s) = \left[\bar{Q}_i(s)\right]_{i=1}^k$\\ 
                \For{each action $a\in \mathpzc{A}_i$ in state, $s$}{
                    Find weighted Q-value, $Q_{p_i}^*(s,a) = w\cdot Q_{p_i}(s, a) + (1-w)(\frac{1}{k}\sum_{i=1}^{k} \bar{Q}_i(s, a))$\\
                    Append $Q_{p_i}^*(s,a)$ to $Q_{p_i}^*(s)$
                }
            Update Q-table with $Q_{p_i}^*(s)$\\
            Choose $a^*\in \mathpzc{A}_i$ for $s$ using $\epsilon$-greedy policy\\
            Execute action, $a^*$, observe $\phi_{p_i}, s'$\\
            Perform $Q_{p_i}(s,a) \leftarrow (1-\alpha)Q_{p_i}(s,a) + \alpha \left[\phi_{p_i}+\Gamma \; \underset{a'}{max}\;Q(s', a')\right]$\\
            Set, $s\leftarrow s'$
            }
            \textbf{If }$\lVert(x_{p_i}, y_{p_i})-(x_{\mathpzc{G}, y_{\mathpzc{G}}})>0\rVert$ \textbf{then}
                continue\\
                \textbf{else} end episode and reset $\mathbb{E}$
        }
    }
    \KwRet{Trained LDP-CMARL model}\\
    \Fn{\LDP{$s, \varepsilon$, b}}{
        \For{$i = 1, 2, ..., k$ advisors}{
            Receive advice request for the state, $s$\\
            With $\varrho_{p_k}$, \For{each action $a\in \mathpzc{A}_i$}{
                find $Q_i(s, a)$ and generate $\eta_i\sim\mathcal{N}(0, b)$
                %\; \forall\; b \geq \frac{\alpha\lvert u-l\rvert}{\varepsilon}$
                \\
                Add LDP-noise,
                $\bar{Q}_i(s, a) = Q_i(s, a)+\eta_i$\\
                \eIf{$\bar{Q}_i(s, a) \notin (l, u)$}{
                    Repeat loop until $\bar{Q}_i(s, a) \in (l, u)$
                }{
                Append $\bar{Q}_i(s, a)$ to $\bar{Q}_i(s)$
                }
            }
            \KwRet{$\bar{Q}_i(s)$}
        }
        \KwRet{$\left[\bar{Q}_i(s) \right]_{i=1}^k$}  
        
    }
    \caption{LDP-CMARL Framework}
    \label{algo:DP_CMARL}
\end{algorithm}
\section{Privacy Exploited Localized Poisoning Attack}
\label{s:problemFormulation}
In this section, we dissect the DP noise exploitation mechanism, formulating adversarial noise profile challenges. We also articulate our threat model and proposed PeLPA algorithm.

\subsection{How can LDP-noise be Exploited for Poisoning Attacks?} \hskip1em \textbf{DP not included.} Considering a non-LDP advising scenario, the agents exchange Q-value knowledge, facilitating learning. We formulate the knowledge as Q-values instead of the recommended actions since the Q-value advising, unlike the action advising, does not impair the performance of the agent’s learning directly \cite{zhu2021q}. Let us assume an anomaly detector at $p_i$'s end that monitors Q-values sequences from advisor agents for all actions in a specific state, $s$. Generally, for a received Q-value, $Q_{p_k}(s)$, from advisor $p_k$, the condition $|Q_{p_{k}}(s)-Q_{0}(s)|\leq \tau$ is consistently maintained, where $\tau$ is a detection threshold and $Q_{0}(s)$, a historical standard Q-value. Any deviation raises an alarm, implying a potential malicious advisor $p_a \in \left[p_k\right]$ with biased Q-values. Nonetheless, to evade detection, the attacker can introduce a bias up to a maximum of $\tau$ relative to the standard value, i.e., $Q_{p_a}(s) \leq Q_{0}(s) + \tau$.

\textbf{DP included.} With an LDP mechanism safeguarding knowledge exchange, any received Q-value, $\bar{Q}{p_k}(s) = Q{p_k}(s) + \eta$, includes noise, $\eta$ following a zero-mean Laplace distribution, $\mathcal{N}(0, b)$, where $b$ is the distribution scale. To prevent false-positive alarms for benign differentially private Q-values, the detector adjusts the previous detection condition to $|\bar{Q}{p_k}(s)-Q_{0}(s)|\leq \tau'$ with $\tau' = \tau \times \kappa; \forall \kappa \in \mathbb{R}$, where $\kappa$ is the tolerance multiplier. This adjustment creates a poisoning window of $\lvert\tau(1-\kappa)\rvert$ that an attacker can exploit, enabling a larger bias in knowledge (i.e., Q-values) without detection. Formally, the attacker shares malicious knowledge, $\bar{Q}{p_a}(s) = Q{p_a}(s)+\eta_a; \forall \eta_a \in \lvert\tau(1-\kappa)\rvert$, where $\eta_a$ denotes the malicious noise drawn from an adversarial noise profile, $\mathcal{N}_a$. Hence, an increase in noise for privacy enhancement also expands the detection and the poisoning window.

\subsection{Challenges in Formulating Adversarial Noise Profile}
\label{noiseProfile} Crafting an adversarial noise profile, $\mathcal{\eta}_a$, that optimizes attack gain while evading anomaly detection poses a technical conundrum. A previous methodology \cite{fang2020local} attempted this by maximizing utility degradation, although this leads to a paradoxical situation in the face of an anomaly detector - more noise aids detection but less noise diminishes the attack gain. A sophisticated alternative, as proposed by \cite{giraldo2020adversarial}, models this as a multi-objective optimization problem, i.e., $\underset{\mathcal{A}}{max}\;\mathpzc{G}(\mathcal{A}, \mathcal{D}) \ni |\bar{Q}_{p_{a}}(s)-Q_{0}(s)|\leq \tau'$ where $\mathcal{A}, \mathcal{D}, \text{and } \mathpzc{G}$ denote the attack, the detect, and the gain function, respectively. The solution of this multi-criteria optimization problem is derived in \cite{giraldo2020adversarial}, where the authors presented an attack impact, $\mu^*_a$, and an optimal adversarial distribution, $\mathcal{N}_a^*(\mu_a^*, b)$ having the probability density function, $f^*_a$ as

\begin{equation}
    f_a^*(x) = \frac{k^2 - b^2}{2bc^2}e^{-\frac{|x-\theta|}{b} + \frac{(x-\theta)}{c}} \;\text{and}\; \mu_a^* = \frac{b^2(\theta-2c)-\theta c^2}{b^2 - c^2}
    \label{eqn:attackDist}
\end{equation}

where $\theta$ is the mean, $b^2$ is the variance, and $c$ is the Lagrange multiplier. $c$ can be solved numerically from \cite{giraldo2020adversarial}: 

\begin{equation}
    \frac{2b^2}{c^2 - b^2} + \ln{(1-\frac{b^2}{c^2})}=\gamma.
    \label{eqn:c}
\end{equation} 

\begin{figure}[!ht]
\centerline{\includegraphics[width=1.0\linewidth]{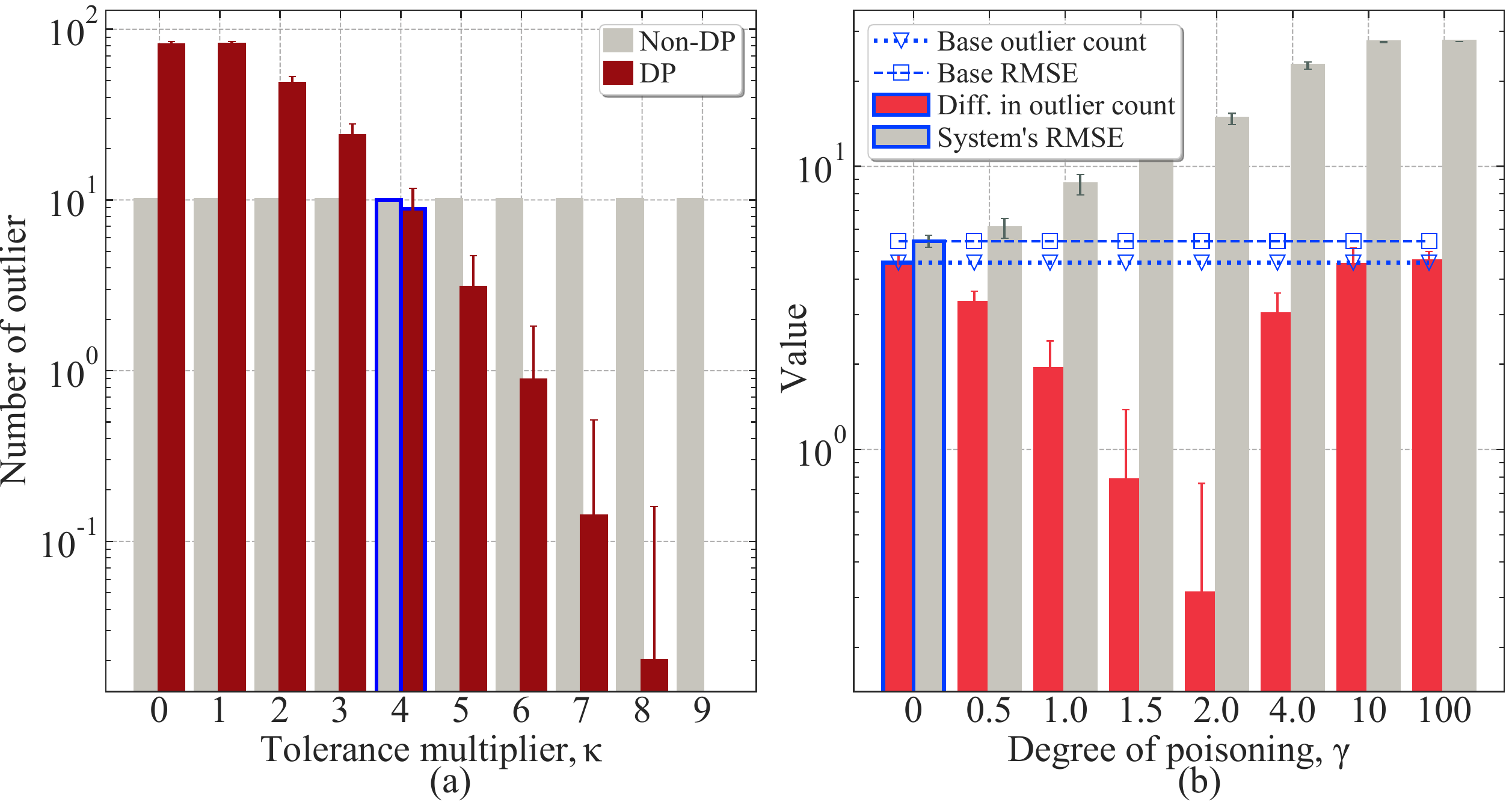}}
\caption{\small (a) Impact of tolerance multiplier, $\kappa$ over detected outliers in both non-DP and DP settings, (b) Impact of degree of knowledge poisoning, $\gamma$ over attack evasion (difference in outlier count between non-attack and attack scenario) and attack gain (System's RMSE).}
\label{fig:outlier_rmse}
\end{figure}

Here, $\gamma$ is the degree of knowledge poisoning; a high $\gamma$ implies a large malicious noise injection (i.e., a higher attack gain) and vice versa. In particular, choosing a high $\gamma$ can lead to unrealistically large Q-values whereas choosing a minuscule $\gamma$ can result in negligible to almost zero attack gain. Consequently, tuning $\gamma$ for an optimal attack is non-trivial but challenging, which, unfortunately, overlooked by literature so far. We address this research gap in section~\ref{s:methodology}. Figure~\ref{fig:outlier_rmse}(a) demonstrates the influence of $\kappa$ and $\gamma$ on detected outliers and RMSE. By adding LDP-noise to $100$ uniform random values, non-DP Q-values detect a steady number of outliers for a fixed $\tau$, whereas LDP implementation significantly increases outlier detection due to benign DP Q-values flagged as false positives. This can be mitigated by setting $\tau' = \tau\times\kappa$. Moreover, an optimal attack approach as per (\ref{eqn:attackDist}) allows successful detection evasion, maintaining the baseline outlier count while inflating the system's RMSE, as shown in Fig.~\ref{fig:outlier_rmse}(b).

\subsection{Attacker's Capability and Knowledge} 
\begin{figure}[!t]
\centerline{\includegraphics[width=\linewidth]
{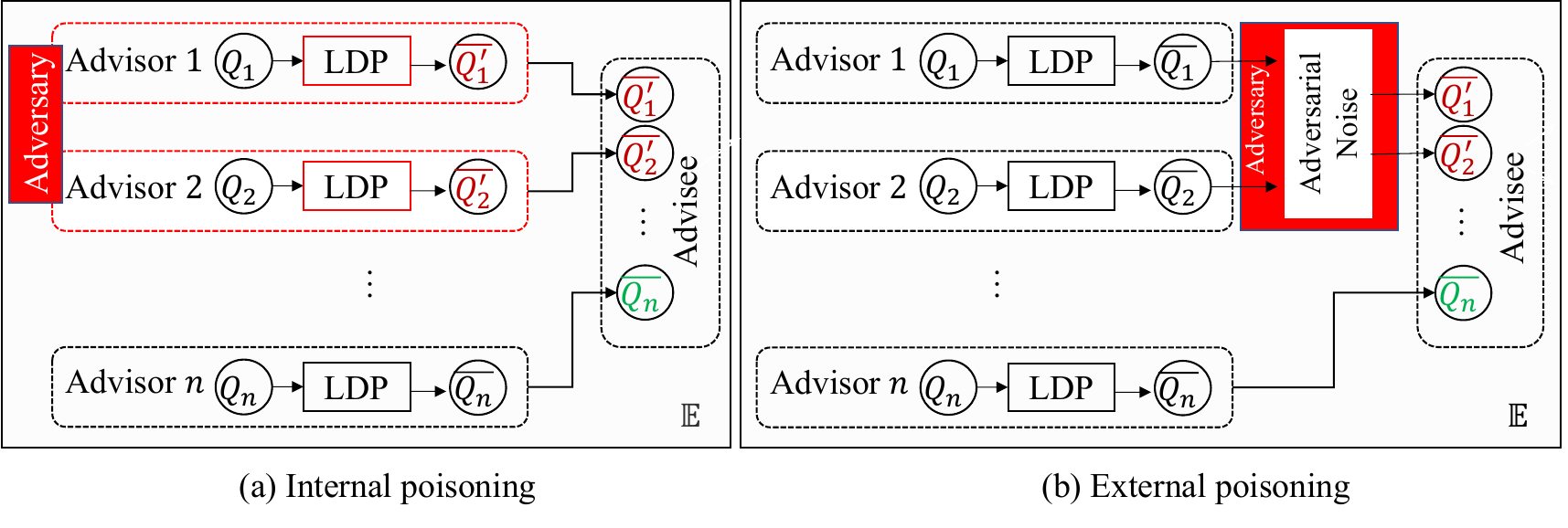}}
\caption{\small (a) Internal poisoning: Attacker compromises advisors and replaces benign LDP process with adversarial LDP process, (b) External poisoning: Attacker compromises the communication path and injects additional malicious noise.} 
\label{fig:poisoning_framework}
\end{figure}

We contemplate an attacker manipulating knowledge submissions to an advisee, either by exploiting susceptible agents (internal threats) or by compromising communication channels (external threats) (Fig.~\ref{fig:poisoning_framework}a, b). The attacker, in line with SOTA research \cite{dwork2019differential}, is presumed to know the publicly available $\varepsilon$-value and noise distribution.
\begin{figure*}[!t]
\centerline{\includegraphics[width=\textwidth]{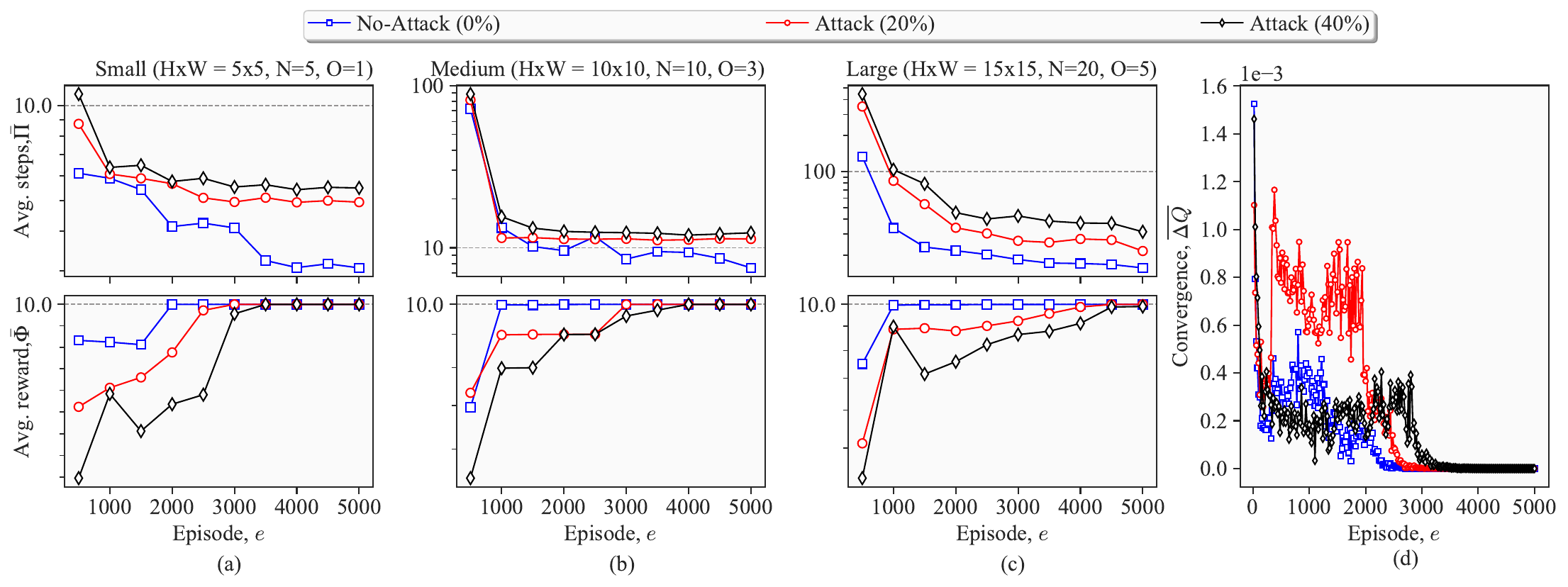}}
\caption{\small Average steps to goal ($\bar{\Pi}$) and obtained reward ($\bar{\Phi}$) analysis for (a) small ($\mathpzc{H}\times\mathpzc{W}=5\times 5, \mathpzc{N}=5, \mathpzc{O}=1$), (b) medium ($\mathpzc{H}\times\mathpzc{W}=10\times 10, \mathpzc{N}=10, \mathpzc{O}=3$), and (c) large-scale ($\mathpzc{H}\times\mathpzc{W}=15\times15, \mathpzc{N}=20, \mathpzc{O}=5$) environments. The number of steps is increased as well as the maximum reward achievement is delayed with more attacks (large attacker ratio). Also, (d) convergence is delayed for both $20\%$ and $40\%$ attacks compared to the no-attack baseline.} 
\label{fig:comparison_convergence}
\end{figure*}
\subsection{Proposed PeLPA Algorithm} \label{s:methodology}
A malevolent advisor, $p_a \in \mathpzc{N}$, could disrupt $p_i$'s convergence by transmitting erroneous information during the knowledge-sharing phase. Having knowledge of $\mathpzc{A, S}, \Phi, (x_{\mathpzc{G}},y_{\mathpzc{G}})$ and $p_i$'s state, $s$, $p_a$ might manipulate larger $Q$-values for a misleading action $a_m$ versus an ideal action $a_h$. This would steer $p_i$ towards a malicious point. Yet, anomalous Q-values could either invite detection or result in an insignificant attack impact. The optimal attack method in section~\ref{noiseProfile} addresses this trade-off. Our proposed PeLPA attack for LDP-CMARL is detailed in Algorithm~\ref{algo:Attack}. $p_a$ continually injects adversarial noises ($\eta_a$) to its Q-values ($Q_a(s,a)$) until    either the malicious Q-values drop below $p_i$'s maximum Q-value for an action $a$, or $\gamma$ exceeds a predetermined poisoning threshold ($\tau_{\gamma}$). Additionally, $p_a$ ensures malicious advice stays within the reward range $\bar{Q}_{p_a}(s, a) \in [l,u]$ to evade detection.

\begin{algorithm}[!ht]
    \SetKwInOut{KwIn}{Input}
    \SetKwInOut{KwOut}{Output}
    \SetKwRepeat{Do}{do}{while}
    \KwIn{$\varepsilon, b, \alpha, Q_{p_i}(s), Q_{p_a}(s)$}
    Initialize $\bar{Q}_{p_a}(s) = \text{[}\;\text{]}$ and set $\gamma\leftarrow 0, \Psi\leftarrow True, \theta\leftarrow 0$\\
    \While{$\Psi$ is True}{
        $\gamma= \gamma+1$\\
        With $b$ and $\gamma$, find $c$ numerically from (\ref{eqn:c})\\
        Then, with $c, \theta$ and $b$, find $\mu^*_{p_a}$ from (\ref{eqn:attackDist})\\
        \For{each $a\in\mathpzc{A}_i$ in state, $s$}{
            \While{$\bar{Q}_{p_a}(s, a) \notin (l, u)$}{
                $\bar{Q}_{p_a}(s,a) = Q_{p_a}(s,a) +\eta_a\sim\mathcal{N}(\mu^*_{p_a}, b)$}
            Append $\bar{Q}_{p_a}(s,a)$ to $\bar{Q}_{p_a}(s)$\\
        }
        $\bar{Q}^*_{p_a}(s) $ =
            $\begin{cases}
                \begin{aligned}
                    & \bar{Q}_{p_a}(s) \;\text{and} \\
                    & \hskip1em\Psi \leftarrow False,
                \end{aligned} & 
                \begin{array}{rr}
                    & \text{if }\bar{Q}_{p_a}(s,a)<Q_{p_i}(s,a) \text{ s.t. }\\
                    &  a \text{ for }{}_{max}Q_{p_i}(s) \text{ or } \gamma>\tau_\gamma
                \end{array}\\
                
                % \bar{Q}_{p_a}(s,a)<Q_{p_i}(s,a) | a \text{ for }{}_{max}Q_{p_i}(s)\\
            Continue\text{,} & \text{Otherwise until } \gamma\leq\tau_\gamma
        \end{cases}$\\

        Set $\bar{Q}_{p_a}(s) = \text{[ ]}$   
    }
    \KwRet{$\bar{Q}^*_{p_a}(s)$}
    
    \caption{Proposed PeLPA Algorithm}
    
    \label{algo:Attack}
    
\end{algorithm}

\section{Experimental Analysis} \label{s:experimentalAnalysis}
In this section, we implement our proposed PeLPA attack in a modified predator-prey domain, following the environmental specifications detailed in section~\ref{subsec:CMARL} \cite{le2017coordinated}. The environment consists of multiple predator agents and one prey. The environment is reset if the initial agent doesn't achieve the goal within a specified number of steps. Table~\ref{tab:parameter} presents the experimental parameters. For comparative insight, we investigate three environment scales: \textbf{small-scale (5x5)}, \textbf{medium-scale (10x10)}, and \textbf{large-scale (15x15)}, exploring $0\%$, $20\%$, and $40\%$ attacker percentages in each. Each experiment is repeated 10 times to average results. We use a privacy budget $\varepsilon = 1.0$ for all results presented, even though a smaller $\varepsilon$ would indicate stronger privacy protection, albeit with larger attack gains.

\textbf{Steps to Goal ($\Pi$) Analysis.} 
The $\bar{\Pi}$-values represent the average steps an agent takes to achieve the goal, with lower values indicating efficient learning. The top three charts of Fig. \ref{fig:comparison_convergence}(a-c) reveals an increase in the required step count to reach the goal as the attacker ratio rises and the environment expands. For example, after $5000$ episodes in a medium-scale environment, $\bar{\Pi} = \{7.52, 11.332, 12.364\}$ for $\{0\%, 20\%, 40\%\}$ attackers, leading to a $\frac{(11.332-7.52)\times 100}{7.52}\approx 50.69\%$ and $\frac{(12.364-7.52)\times 100}{7.52}\approx 64.41\%$ increase in average \textit{steps to goal} for $20\%$ and $40\%$ attackers, respectively.

\textbf{Reward ($\Phi$) Analysis.} Similarly, the $\bar{\Phi}$-values represent average rewards obtained by agents as shown in the bottom three charts of Fig. \ref{fig:comparison_convergence}(a-c). Our experiments exhibit a decrease in the speed of obtaining optimal rewards as the attacker ratio escalates. For instance, in a medium-scale environment, $\{2500, 3500, 4000\}$ episodes are requisite to attain the optimal $\bar{\Phi}$, for $\{0\%, 20\%, 40\%\}$ attackers, respectively. This leads to a $\frac{3500}{2500} \approx 1.4$x and $\frac{4000}{2500} \approx 1.6$x time increase in optimal $\bar{\Phi}$ acquisition for $20\%$ and $40\%$ attackers, respectively.
\setlength{\textfloatsep}{8pt}% 
\begin{table}[!ht]
\centering
\caption{\small Parameter value. $\alpha$: learning rate, $\epsilon$: exploration-exploitation probability, $\Gamma$: discount factor, $B$: communication budget, $w$: aggregation factor, $\tau, \tau', \tau_{\gamma}$: predefined threshold, $\phi$: reward, $\varepsilon$: privacy budget.}
\label{tab:parameter}
\begin{adjustbox}{max width=\linewidth}
\begin{tabular}{c|c|c|c|c|c|c|c}
\toprule
Parameter &
$\alpha$                                                      & $\epsilon$                                                    & $\Gamma$                                                      & \multicolumn{1}{c|}{$B^{tot}_{p_i}$}           & $B^{tot}_{p_a}$                          & $w$                                          & $\tau_{\gamma}$   \\ 
\midrule
Value &
0.10                                                                           & 0.08                                                                           & 0.80                                                                           & 100,000                           & 10,000                                              & 0.90                                           & 12                       \\
\midrule
\multicolumn{1}{c|}{Parameter} &
\multicolumn{1}{c|}{$\phi_{\mathpzc{G}}$} & \multicolumn{1}{c|}{$\phi_{\mathpzc{F}}$} & \multicolumn{1}{c|}{$\phi_{\mathpzc{O}}$} & $\phi_{\mathpzc{W}}$ & \multicolumn{1}{l|}{$\varepsilon$} & \multicolumn{1}{c|}{$\tau$} & $\tau'$  \\ 
\midrule
\multicolumn{1}{c|}{Value}  & 
\multicolumn{1}{c|}{10.0}                                                      & \multicolumn{1}{l|}{0.50}                                                      & \multicolumn{1}{c|}{-1.50}                                                     & -0.50                                                     & \multicolumn{1}{l|}{1.0}                            & \multicolumn{1}{l|}{100}                      & 100,000   \\
\bottomrule
\end{tabular}
\end{adjustbox}
\end{table}

\textbf{Convergence ($\Delta Q$) Analysis.} 
To gauge the effectiveness of our proposed attack, we conduct a convergence analysis based on $\overline{\Delta Q}$ values, i.e., the average of the deviation of Q-values from the optimal value ($Q^*$). An optimal learning process would have $\overline{\Delta Q}$ values tending to zero, and our analysis confirms this behavior is impeded as the attacker ratio increases. This delay in convergence correlates with the increase in attacker prevalence. Specifically, in a medium-scale environment, $\overline{\Delta Q}$ falls below $10e^{-6}$ following $\{2360, 2800, 3280\}$ episodes for $\{0\%, 20\%, 40\%\}$ attackers. Consequently, convergence is delayed by $\frac{2800}{2360}\approx 1.18$x and $\frac{3280}{2360}\approx 1.38$x for attacker ratios of $20\%$ and $40\%$, respectively.

\begin{figure}[!t]
\centerline{\includegraphics[width=\linewidth]{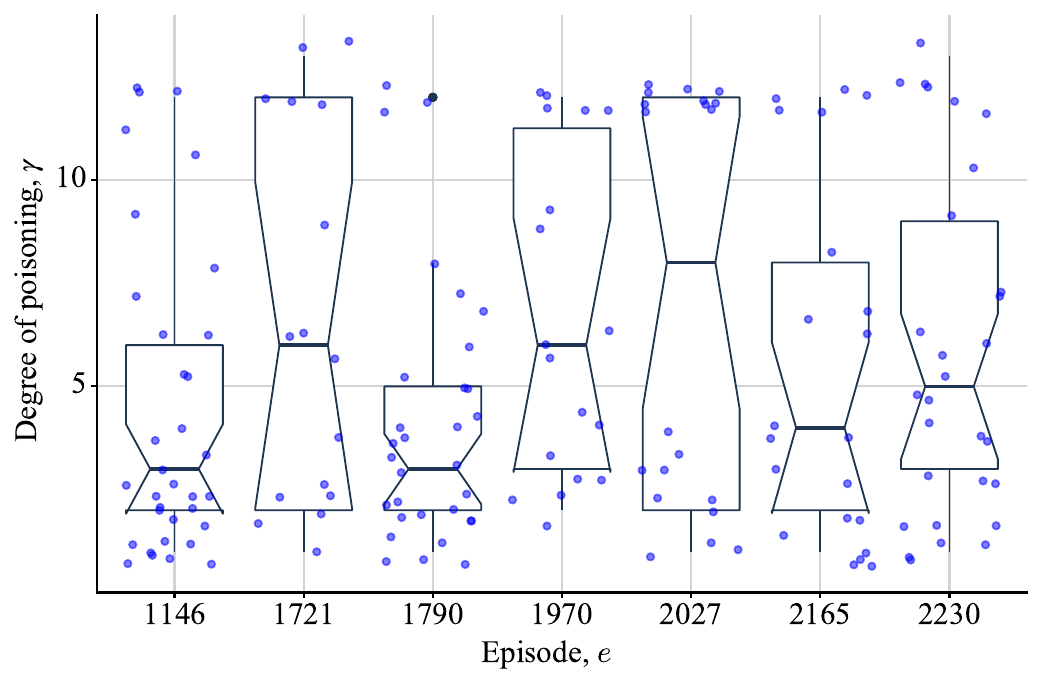}}
\caption{\small Distribution of the degree of knowledge poisoning, ($\gamma$) in some example episodes. For instance, in episode $1146$, the attacker maintained the $\gamma$ value under $5$ for most of the steps but increased it to more than $10$ for a few steps.}
\label{fig:degree}
\end{figure}

\textbf{Adaptive Degree of Knowledge Poisoning ($\gamma$).}
\label{degreeOfPoisoning}
Finally, we consider the degree of knowledge poisoning, $\gamma$, demonstrating its distribution and symmetry in various scenarios as shown in Fig.~\ref{fig:degree}. This parameter is adjusted following line $10$ in Algorithm~\ref{algo:Attack}, showing varied instances of its manipulation across different episodes. We only present the episodes in which the attacker adjusted the $\gamma$ value more than $20$ times. For example, in episode $1146$, the attacker maintained the $\gamma$ value under $5$ for most of the steps but increased it to more than $10$ for a few steps. Contrarily, in episode $2027$, the attacker never sets $\gamma$ in the range of $\left[5,10\right]$.

\section{Conclusions} \label{s:Conclusion}

This paper highlights the potential security risks of using DP in CMARL algorithms and proposes a new adaptive and localized knowledge poisoning attack technique (PeLPA) to exploit DP-noise and prevent optimal convergence of the CMARL model. The proposed PeLPA technique is designed to evade SOTA anomaly detection techniques and degrade the multiagent learning performance. The effectiveness of the proposed attack technique is demonstrated through extensive experimental analysis in varying environment scales. The study fills a research gap in the literature and sheds light on the need for stronger security measures in LDP-CMARL systems.

\bibliography{reference}

\begin{thebibliography}{10}
\providecommand{\url}[1]{#1}
\csname url@samestyle\endcsname
\providecommand{\newblock}{\relax}
\providecommand{\bibinfo}[2]{#2}
\providecommand{\BIBentrySTDinterwordspacing}{\spaceskip=0pt\relax}
\providecommand{\BIBentryALTinterwordstretchfactor}{4}
\providecommand{\BIBentryALTinterwordspacing}{\spaceskip=\fontdimen2\font plus
\BIBentryALTinterwordstretchfactor\fontdimen3\font minus
  \fontdimen4\font\relax}
\providecommand{\BIBforeignlanguage}[2]{{%
\expandafter\ifx\csname l@#1\endcsname\relax
\typeout{** WARNING: IEEEtran.bst: No hyphenation pattern has been}%
\typeout{** loaded for the language `#1'. Using the pattern for}%
\typeout{** the default language instead.}%
\else
\language=\csname l@#1\endcsname
\fi
#2}}
\providecommand{\BIBdecl}{\relax}
\BIBdecl

\bibitem{Zhou2022}
W.~Zhou, D.~Chen, J.~Yan, Z.~Li, H.~Yin, and W.~Ge, ``Multi-agent reinforcement
  learning for cooperative lane changing of connected and autonomous vehicles
  in mixed traffic,'' \emph{Autonomous Intelligent Systems}, vol.~2, 12 2022.

\bibitem{da2017simultaneously}
F.~L. Da~Silva, R.~Glatt, and A.~H.~R. Costa, ``Simultaneously learning and
  advising in multiagent reinforcement learning,'' in \emph{Proceedings of the
  16th conference on autonomous agents and multiagent systems}, ser. AAMAS '17,
  2017, pp. 1100--1108.

\bibitem{le2017coordinated}
H.~M. Le, Y.~Yue, P.~Carr, and P.~Lucey, ``Coordinated multi-agent imitation
  learning,'' in \emph{International Conference on Machine Learning}.\hskip 1em
  plus 0.5em minus 0.4em\relax PMLR, 2017, pp. 1995--2003.

\bibitem{li2022privacy}
Q.~Li, B.~Guo, and Z.~Wang, ``A privacy-preserving multi-agent updating
  framework for self-adaptive tree model,'' \emph{Peer-to-Peer Networking and
  Applications}, vol.~15, no.~2, pp. 921--933, 2022.

\bibitem{zou2020privacy}
Y.~Zou, Z.~Zhang, M.~Backes, and Y.~Zhang, ``Privacy analysis of deep learning
  in the wild: Membership inference attacks against transfer learning,''
  \emph{arXiv preprint arXiv:2009.04872}, 2020.

\bibitem{dwork2006}
C.~Dwork, ``Differential privacy,'' in \emph{Automata, Languages and
  Programming}, M.~Bugliesi, B.~Preneel, V.~Sassone, and I.~Wegener, Eds.\hskip
  1em plus 0.5em minus 0.4em\relax Berlin, Heidelberg: Springer Berlin
  Heidelberg, 2006, pp. 1--12.

\bibitem{ye2022differential}
D.~Ye, T.~Zhu, Z.~Cheng, W.~Zhou, and P.~S. Yu, ``Differential advising in
  multiagent reinforcement learning,'' \emph{IEEE Transactions on Cybernetics},
  vol.~52, no.~6, pp. 5508--5521, 2022.

\bibitem{Abahussein2023}
S.~Abahussein, T.~Zhu, D.~Ye, Z.~Cheng, and W.~Zhou, ``Protect trajectory
  privacy in food delivery with differential privacy and multi-agent
  reinforcement learning,'' in \emph{Advanced Information Networking and
  Applications}, L.~Barolli, Ed.\hskip 1em plus 0.5em minus 0.4em\relax Cham:
  Springer International Publishing, 2023, pp. 48--59.

\bibitem{ye2022one}
D.~Ye, S.~Shen, T.~Zhu, B.~Liu, and W.~Zhou, ``One parameter
  defense—defending against data inference attacks via differential
  privacy,'' \emph{IEEE Transactions on Information Forensics and Security},
  vol.~17, pp. 1466--1480, 2022.

\bibitem{scheikl2021cooperative}
P.~M. Scheikl, B.~Gyenes, T.~Davitashvili, R.~Younis, A.~Schulze, B.~P.
  M{\"u}ller-Stich, G.~Neumann, M.~Wagner, and F.~Mathis-Ullrich, ``Cooperative
  assistance in robotic surgery through multi-agent reinforcement learning,''
  in \emph{2021 IEEE/RSJ International Conference on Intelligent Robots and
  Systems (IROS)}.\hskip 1em plus 0.5em minus 0.4em\relax IEEE, 2021, pp.
  1859--1864.

\bibitem{la2014multirobot}
H.~M. La, R.~Lim, and W.~Sheng, ``Multirobot cooperative learning for predator
  avoidance,'' \emph{IEEE Transactions on Control Systems Technology}, vol.~23,
  no.~1, pp. 52--63, 2014.

\bibitem{prasad2019multi}
A.~Prasad and I.~Dusparic, ``Multi-agent deep reinforcement learning for zero
  energy communities,'' in \emph{2019 IEEE PES innovative smart grid
  technologies Europe (ISGT-Europe)}.\hskip 1em plus 0.5em minus 0.4em\relax
  IEEE, 2019, pp. 1--5.

\bibitem{kiran2021deep}
B.~R. Kiran, I.~Sobh, V.~Talpaert, P.~Mannion, A.~A.~A. Sallab, S.~Yogamani,
  and P.~Pérez, ``Deep reinforcement learning for autonomous driving: A
  survey,'' \emph{IEEE Transactions on Intelligent Transportation Systems},
  vol.~23, no.~6, pp. 4909--4926, 2022.

\bibitem{cao2021data}
X.~Cao, J.~Jia, and N.~Z. Gong, ``Data poisoning attacks to local differential
  privacy protocols,'' in \emph{30th $\{$USENIX$\}$ Security Symposium
  ($\{$USENIX$\}$ Security 21)}, 2021.

\bibitem{fang2020local}
M.~Fang, X.~Cao, J.~Jia, and N.~Gong, ``Local model poisoning attacks to
  byzantine-robust federated learning,'' in \emph{29th $\{$USENIX$\}$ Security
  Symposium ($\{$USENIX$\}$ Security 20)}, 2020, pp. 1605--1622.

\bibitem{mohammadi2023implicit}
M.~Mohammadi, J.~N{\"o}ther, D.~Mandal, A.~Singla, and G.~Radanovic, ``Implicit
  poisoning attacks in two-agent reinforcement learning: Adversarial policies
  for training-time attacks,'' \emph{arXiv preprint arXiv:2302.13851}, 2023.

\bibitem{cheu2021manipulation}
A.~Cheu, A.~Smith, and J.~Ullman, ``Manipulation attacks in local differential
  privacy,'' in \emph{2021 IEEE Symposium on Security and Privacy (SP)}.\hskip
  1em plus 0.5em minus 0.4em\relax IEEE, 2021, pp. 883--900.

\bibitem{Hossain2021Desmp}
M.~T. Hossain, S.~Islam, S.~Badsha, and H.~Shen, ``Desmp: Differential
  privacy-exploited stealthy model poisoning attacks in federated learning,''
  in \emph{2021 17th International Conference on Mobility, Sensing and
  Networking (MSN)}, 2021, pp. 167--174.

\bibitem{giraldo2020adversarial}
J.~Giraldo, A.~Cardenas, M.~Kantarcioglu, and J.~Katz, ``Adversarial
  classification under differential privacy,'' in \emph{Network and Distributed
  Systems Security (NDSS) Symposium 2020}, 2020.

\bibitem{hossain2022adversarial}
M.~T. Hossain, S.~Badsha, H.~La, H.~Shen, S.~Islam, I.~Khalil, and X.~Yi,
  ``Adversarial analysis of the differentially-private federated learning in
  cyber-physical critical infrastructures,'' 2022.

\bibitem{wang2019privacy}
B.~Wang and N.~Hegde, ``Privacy-preserving q-learning with functional noise in
  continuous spaces,'' \emph{Advances in Neural Information Processing
  Systems}, vol.~32, 2019.

\bibitem{Wei2022}
D.~Wei, J.~Zhang, M.~Shojafar, S.~Kumari, N.~Xi, and J.~Ma, ``Privacy-aware
  multiagent deep reinforcement learning for task offloading in vanet,''
  \emph{IEEE Transactions on Intelligent Transportation Systems}, pp. 1--15,
  2022.

\bibitem{figura2021adversarial}
M.~Figura, K.~C. Kosaraju, and V.~Gupta, ``Adversarial attacks in
  consensus-based multi-agent reinforcement learning,'' in \emph{2021 American
  Control Conference (ACC)}.\hskip 1em plus 0.5em minus 0.4em\relax IEEE, 2021,
  pp. 3050--3055.

\bibitem{Xie2022}
Z.~Xie, Y.~Xiang, Y.~Li, S.~Zhao, E.~Tong, W.~Niu, J.~Liu, and J.~Wang,
  ``Security analysis of poisoning attacks against multi-agent reinforcement
  learning,'' in \emph{Algorithms and Architectures for Parallel Processing},
  Y.~Lai, T.~Wang, M.~Jiang, G.~Xu, W.~Liang, and A.~Castiglione, Eds.\hskip
  1em plus 0.5em minus 0.4em\relax Cham: Springer International Publishing,
  2022, pp. 660--675.

\bibitem{hossain2023BRNES}
M.~T. Hossain, H.~M. La, and S.~Badsha, ``Brnes: Enabling security and
  privacy-aware experience sharing in multiagent robotic and autonomous systems
  [manuscript submitted for publication],'' \emph{{Department of Computer
  Science and Engineering, University of Nevada, Reno}}, 2023.

\bibitem{giraldo2017security_2}
J.~Giraldo, A.~A. Cardenas, and M.~Kantarcioglu, ``Security vs. privacy: How
  integrity attacks can be masked by the noise of differential privacy,'' in
  \emph{2017 American Control Conference (ACC)}.\hskip 1em plus 0.5em minus
  0.4em\relax IEEE, 2017, pp. 1679--1684.

\bibitem{hossain2021privacy}
M.~T. Hossain, S.~Badsha, and H.~Shen, ``Privacy, security, and utility
  analysis of differentially private cpes data,'' in \emph{2021 IEEE Conference
  on Communications and Network Security (CNS)}.\hskip 1em plus 0.5em minus
  0.4em\relax IEEE, 2021, pp. 65--73.

\bibitem{Neera2023}
J.~Neera, X.~Chen, N.~Aslam, K.~Wang, and Z.~Shu, ``Private and utility
  enhanced recommendations with local differential privacy and gaussian mixture
  model,'' \emph{IEEE Transactions on Knowledge and Data Engineering}, vol.~35,
  no.~4, pp. 4151--4163, 2023.

\bibitem{wang2020comprehensive}
T.~Wang, X.~Zhang, J.~Feng, and X.~Yang, ``A comprehensive survey on local
  differential privacy toward data statistics and analysis,'' \emph{Sensors},
  vol.~20, no.~24, p. 7030, 2020.

\bibitem{zhu2021q}
C.~Zhu, H.-F. Leung, S.~Hu, and Y.~Cai, ``A q-values sharing framework for
  multi-agent reinforcement learning under budget constraint,'' \emph{ACM
  Transactions on Autonomous and Adaptive Systems (TAAS)}, vol.~15, no.~2, pp.
  1--28, 2021.

\bibitem{dwork2019differential}
C.~Dwork, N.~Kohli, and D.~Mulligan, ``Differential privacy in practice: Expose
  your epsilons!'' \emph{Journal of Privacy and Confidentiality}, vol.~9,
  no.~2, 2019.

\end{thebibliography}
\nocite{*}
\bibliographystyle{IEEEtran}

\end{document}